\documentclass[preprint]{JASA-EL}

\usepackage{amsmath} 
\usepackage{gensymb} 
\usepackage{tipa}

\begin{document}


\title[Scaling laws for nonlinear dynamical models]{Scaling laws for nonlinear dynamical models of articulatory control}

\author{Sam Kirkham}
\email{s.kirkham@lancaster.ac.uk}
\affiliation{Phonetics Laboratory, Lancaster University}



\date{\today} 

\begin{abstract}
Dynamical theories of speech use computational models of articulatory control to generate quantitative predictions and advance understanding of speech dynamics. The addition of a nonlinear restoring force to task dynamic models is a significant improvement over linear models, but nonlinearity introduces challenges with parameterization and interpretability. We illustrate these problems through numerical simulations and introduce solutions in the form of scaling laws. We apply the scaling laws to a cubic model and show how they facilitate interpretable simulations of articulatory dynamics, and can be theoretically interpreted as imposing physical and cognitive constraints on models of speech movement dynamics.
\end{abstract}


\maketitle



\section{Introduction}

The task dynamic model of speech production is a theoretical and mathematical model of how movement goals are controlled in speech \citep{fowler1980, browman-goldstein1992, saltzman-munhall1989, iskarous2017}. The standard model of task dynamics uses the critically damped harmonic oscillator in (\ref{sm89}) as a model of the articulatory gesture, where $x$ is the position of the system, $\dot{x}$ is velocity, $\ddot{x}$ is acceleration, $m$ is mass, $b$ is a damping coefficient, $k$ is a stiffness coefficient, and $T$ is the target or equilibrium position (see \citealt{iskarous2017} for a tutorial introduction). The initial conditions are defined as $x(0) = x_{0}$ and $\dot{x}(0) = \dot{x_{0}}$. The damping coefficient $b$ in a critically damped harmonic oscillator is defined as $b = 2\sqrt{mk}$, where $m = 1$ in most formulations, but see \citet{simko-cummins2010} for an embodied task dynamics where dynamics are defined over physical masses.

\begin{equation}
m\ddot{x} + b\dot{x} + k(x - T) = 0
\label{sm89}
\end{equation}

The linear dynamical model fails to predict characteristics of empirical velocity trajectories, because it significantly underestimates time-to-peak velocity with unrealistically early and narrow velocity peaks compared with those seen in empirical data \citep{byrd-saltzman1998}. One solution, which forms the subject of the current study, is the addition of a nonlinear restoring force \citep{sorensen-gafos2016}, such as the term  $dx^3$ in Equation (\ref{sg16}).

\begin{equation}
m\ddot{x} + b\dot{x} + k(x - T) - d(x - T)^3 = 0
\label{sg16}
\end{equation}

\begin{figure}[h]
\includegraphics[width=7.5cm]{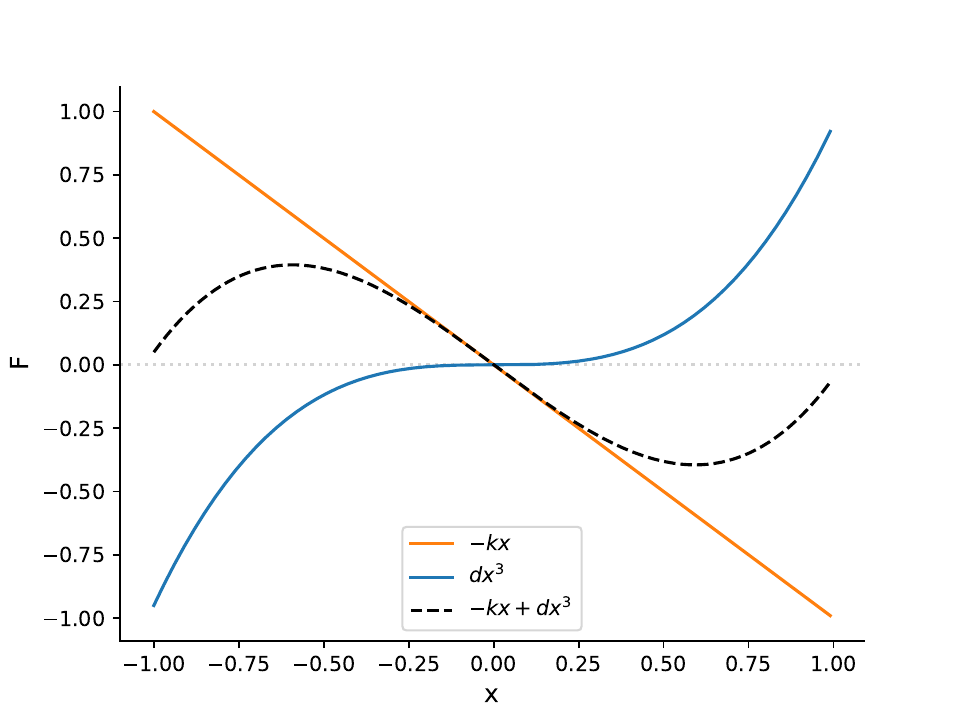}
\includegraphics[width=7.5cm]{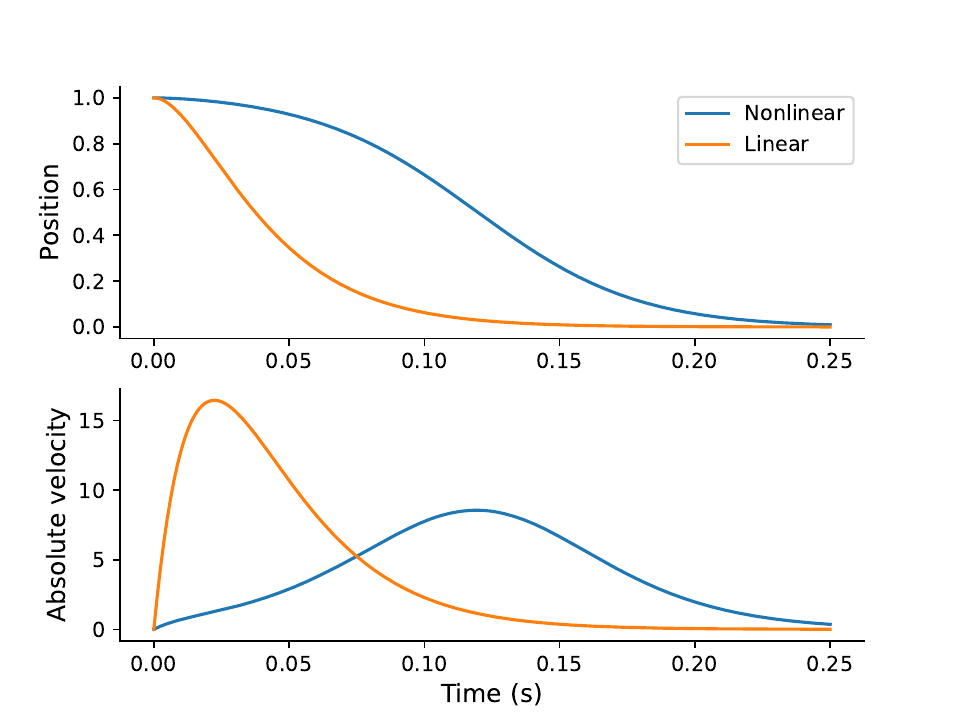}
\includegraphics[width=7.5cm]{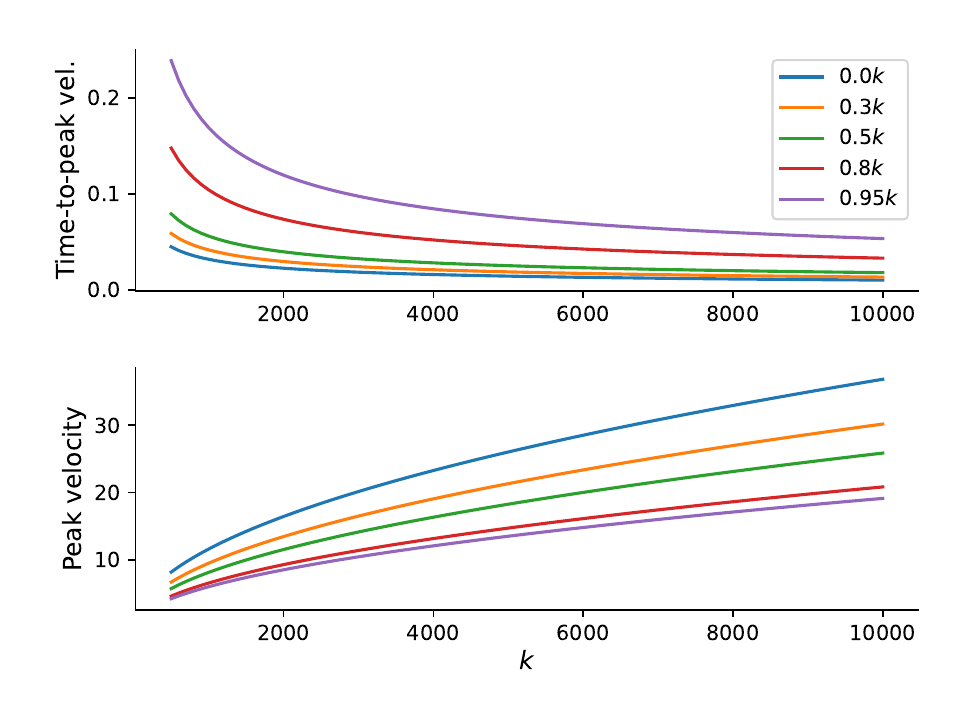}
\includegraphics[width=7.5cm]{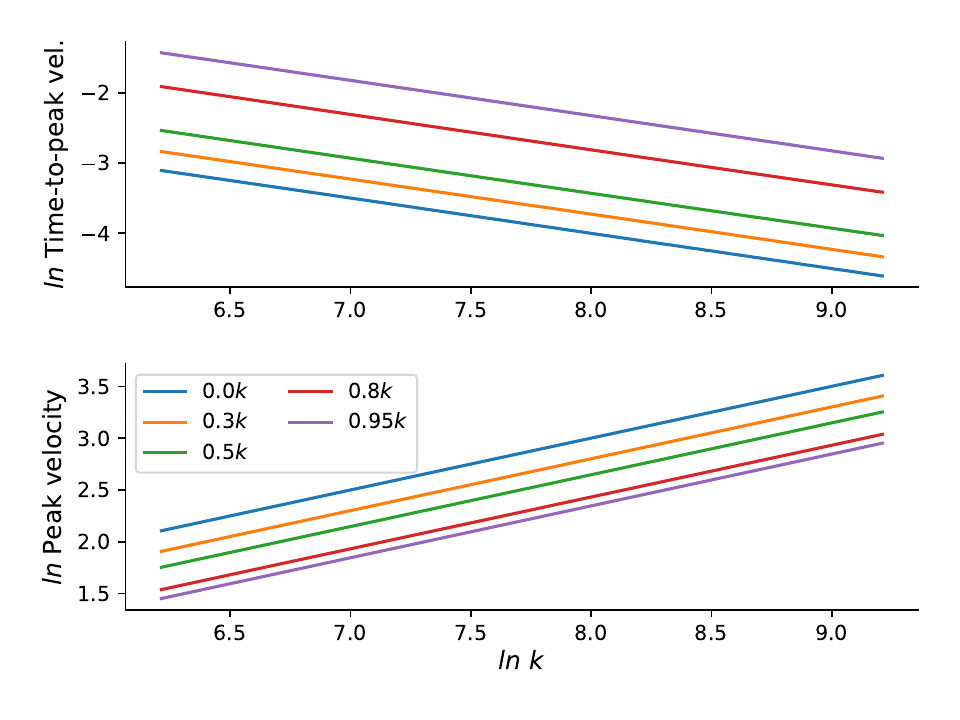}
\caption{TOP LEFT: Stiffness functions of the linear, cubic and summed restoring forces, where $k = 1$ and $F$ refers to the forces specified in the legend as a function of $x$. TOP RIGHT: A comparison of position and velocity trajectories generated by the linear ($d = 0$) and nonlinear ($d = 0.95$) models, where $x_{0} = 1, \dot{x}_{0} = 1, T = 0, k = 2000$. BOTTOM LEFT: Power function of $k$ against time-to-peak velocity (top) and peak velocity (bottom). BOTTOM RIGHT: Power function of the natural logarithms of $k$ against time-to-peak velocity (top) and peak velocity (bottom).}
\label{models}
\end{figure}

The left panel in Figure \ref{models} visualizes the linear $-kx$ and cubic $dx^3$ restoring forces, in addition to the sum of linear and cubic forces $-kx + dx^3$.\footnote{All simulations were implemented in Python. Differential equations were solved using an Explicit Runge-Kutta method of order 5(4) with $\Delta t  = 0.001$ via SciPy's \texttt{integrate.solve\_ivp} function \citep{SciPy-NMeth2020}. The stiffness parameter $k$ is defined as $2/\Delta t$ unless otherwise noted, with $b = 2\sqrt{mk}$.} The right panel in Figure \ref{models} shows a comparison between time-varying position and velocity trajectories generated by the linear and nonlinear models, with identical parameters except $d$ ($k = 2000$, $x_{0} = 1$, $\dot{x}_{0} = 0$, $T = 0$). A value of $d = 0$ is equivalent to a linear model that cancels out the $dx^3$ term, thus representing the linear model, while $d = 0.95k$ produces a quasi-symmetrical velocity shape under these specific conditions.\footnote{Note that $d$ is defined as a multiple of $k$ in order to achieve appropriately scaling between the linear and nonlinear forces.}

A symmetrical velocity trajectory is outside the scope of the standard linear model in Equation (\ref{sm89}), but the use of a nonlinear model is not the only solution. The first approach is the use of a different activation function. In \citet{saltzman-munhall1989} gestural activation is governed by an on/off step function, with instantaneous changes in the target value. \citet{byrd-saltzman1998} instead propose ramped activation, where the parameters of the dynamical system explicitly depend on time, allowing for empirically-realistic time-to-peak velocity. A further development is the use of arbitrary gestural activation functions, which can be learned from data \citep{tilsen2020}. It must be stressed that the idea of continuous gestural activation is fundamentally different from the \citet{sorensen-gafos2016} model, which retains step function gestural activation and instead achieves appropriate velocity characteristics via intrinsic nonlinear gestural dynamics. The distinction here is between autonomous dynamics during the period in which gestural activation is constant (as in \citealt{saltzman-munhall1989, sorensen-gafos2016}) versus non-autonomous dynamics during activation with time-varying parameter values (as in \citealt{byrd-saltzman1998, tilsen2020}). A second approach is to relax the critical damping constraint entirely and recast the gesture as an under-damped half-cycle linear oscillator \citep{kirkham2024}. This improves on the standard linear model in generating symmetrical velocity trajectories and appropriate time-to-peak velocities, but introduces the need for an extrinsic mechanism to avoid target overshoot and unintended oscillation.

The aim of the present study is to explore the numerical parameterization of the nonlinear term in the \citet{sorensen-gafos2016} model specifically, as well as in nonlinear task dynamic models more generally. One issue that we address below is that the effect of any nonlinear term, such as $dx^3$, inherently depends on the distance between the initial position and the target. While an inherent feature of such models, this presents some practical considerations when (i) simulating similar velocity trajectories across articulators or tract variables with varying movement distances; (ii) achieving numerical stability when fitting the model to empirical data; (iii) interpreting parameter values when estimated from empirical data. We first illustrate the problem and then introduce simple numerical methods for examining the relation between nonlinearity and movement distance. We offer two simple solutions based on the same idea: local normalization around an equilibrium point relative to initial position, and global normalization relative to the potential movement range for a given articulator or tract variable. Python code is provided for reproducing all simulations in this article at: \url{https://osf.io/nrxz5/?view_only=e514f671740e43248c230ac6ab35a347} (to be replaced with public link upon acceptance).

\section{Parameters in nonlinear dynamical models}

\subsection{Stiffness and temporal variation}

Before outlining the mechanics of the nonlinear term in the \citet{sorensen-gafos2016} model, we first illustrate the behaviour of the other parameters, which is important for understanding the nonlinear forces. To re-cap, the model is:

\begin{equation}
m\ddot{x} + b\dot{x} + k(x-T) - d(x-T)^3 = 0
\label{sg16_2}
\end{equation}

where $m = 1$ and $b = 2\sqrt k$ in critically damped versions of the model. As a result, we focus on the effects of $k$ on movement characteristics and how it interacts with $d$. The stiffness coefficient $k$ governs the strength of the restoring force; in other words, how quickly the system reaches its equilibrium position. Higher stiffness values result in faster time-to-peak velocity, where the relationship between $k$ and time-to-peak velocity follows a power law $\alpha k^{-\frac{1}{2}}$, with $\alpha$ being larger for larger values of $d$. For example, when $d = 0k$, $\alpha = 1$ and when $d = 0.95k$, $\alpha = 5.4$. The qualitative relationship between stiffness and time-to-peak velocity is the same across different values of $d$, such that the effects of $k$ on time-to-peak velocity follow the same law irrespective of the value of $d$, but the specific quantitative values do vary for the same value of $k$ across different values of $d$. The same is true of the relationship between $k$ and the amplitude of peak velocity, which follows the power law $\alpha k^{\frac{1}{2}}$, where $\alpha$ scales inversely with the value of $d$. For example, when $d = 0k$, $\alpha = 0.37$ and for $d = 0.95k$, $\alpha = 0.19$. These relations are visualized in the bottom left of Figure \ref{models}, which shows the effect of variation in $k$ on peak velocity and time-to-peak velocity at five selected values of $d$, where $x_{0} = 1$, $T = 0$. The bottom right panel shows the natural logarithms of the same variables, with a linear relationship in the log-log plot indicating a power law.

\subsection{Nonlinear cubic term}
\label{cubic-term}

\citet{sorensen-gafos2016} introduced the nonlinear cubic term $dx^3$ in order to make the strength of the restoring force nonlinearly dependent on movement distance. This is what allows for quasi-symmetrical velocity trajectories when $d \approx 0.95k$. In this model, the linear $kx$ and nonlinear $dx^3$ terms are proportionally scaled as in (\ref{dk}). When the absolute movement distance between the initial position and target $|x_{0 } - T| = 1$, $d = 0.95$ will produce a quasi-symmetrical velocity trajectory.

\begin{equation}
d' = dk
\label{dk}
\end{equation}

\begin{figure}[h]
\includegraphics[width=7.5cm]{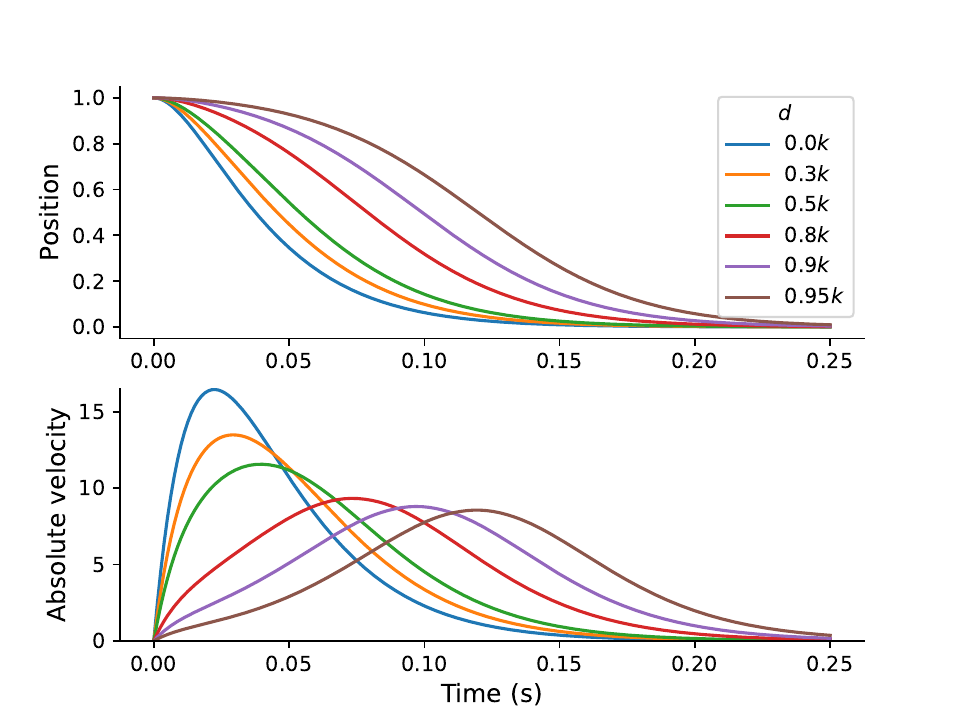}
\includegraphics[width=7.5cm]{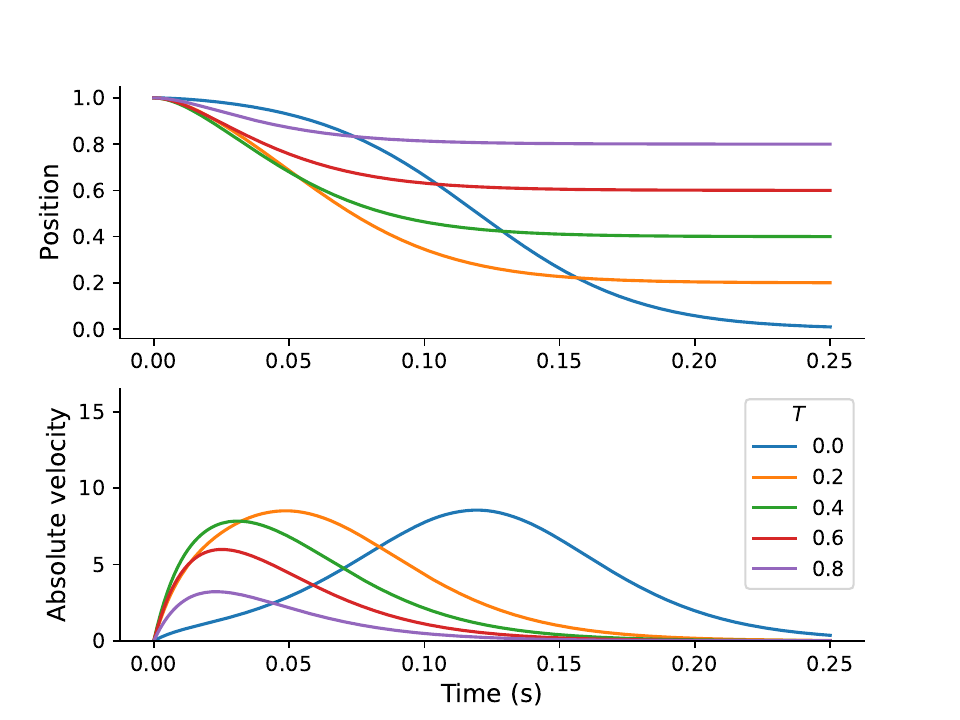}
\includegraphics[width=7.5cm]{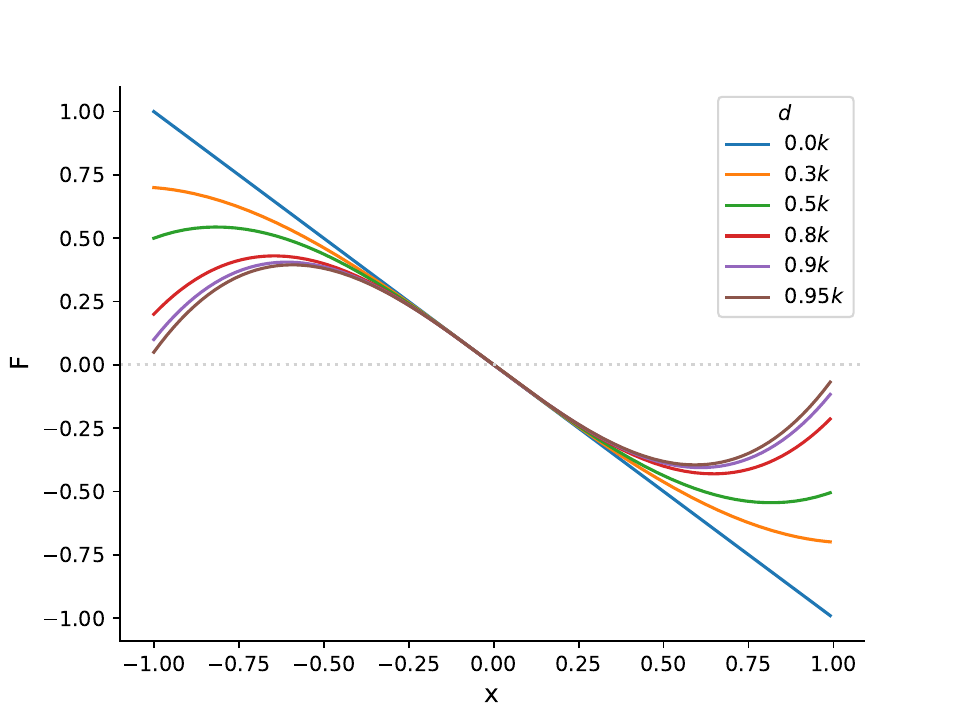}
\includegraphics[width=7.5cm]{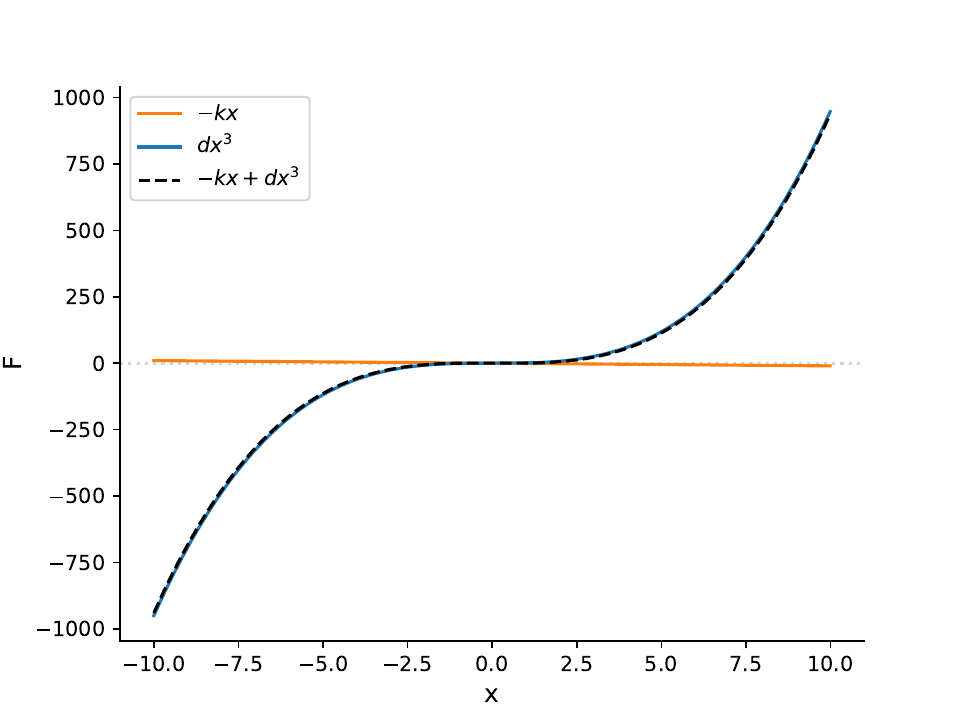}
\caption{TOP LEFT: Simulated position and velocity trajectories, with $x_{0} = 1$, $\dot{x}_{0} = 0$, $k = 2000$, $T = 0.0$ with varying values of $d$; and TOP RIGHT: The same simulations but across varying values of $T$, where $d = 0.95k, k = 2000$. BOTTOM LEFT: Nonlinear restoring force $-kx + dx^3$ ($k = 1$) for values of $d$ corresponding to top left plot, where $F$ refers to the forces specified in the legend as a function of $x$. BOTTOM RIGHT: The restoring forces for $d = 0.95k$ over the range $[-10, 10]$ without scaling.}
\label{fig:d}
\end{figure}

Figure \ref{fig:d} (left) shows that for $|x_{0 } - T| = 1$ then $d = 0.95k$ produces a symmetrical velocity profile, while lower values of $d$ result in earlier time-to-peak velocities and higher peak velocity. This is exactly the scenario described by \citet{sorensen-gafos2016}. When $|x_{0 } - T| \neq1$ the same value of $d$ will produce differently shaped velocity trajectories for different movement distances, which increasingly diverge as $|x_{0 } - T|$ gets further from 1. Figure \ref{fig:d} (right) shows this via simulations ($x_{0} = 1$, $\dot{x}_{0} = 0$, $k = 2000$, $d = 0.95k$) where the target varies across $T = \{0.0, 0.2, ..., 0.8\}$. As movement distance decreases, time-to-peak velocity decreases \textit{and} velocity amplitude decreases nonlinearly. The model can, therefore, capture observed nonlinear relations between movement distance and time-to-peak velocity \citep{munhall-etal1985, ostry-etal1987}, as described by \citet{sorensen-gafos2016}.

A numerical problem with the parameterization of the nonlinear term arises when the movement distance is greater than $|1|$. For example, $d = 0.95k$ when $|x_{0} - T| > 1$ quickly becomes numerically unstable, as the cubic term produces increasingly large values when $dkx^3 > k$. For this reason, the same value of $d$ does not produce the same effects across different movement scales. The differential effects of the same $d$ value across different movement ranges is illustrated in the restoring forces in Figure \ref{fig:d} (bottom right) over a range of $[-10, 10]$. Once the cubic term acts on values above $|1|$ the resulting solution quickly goes to extreme values that are not physically possible for gestural systems. In this case, the $dx^3$ and $-kx + dx^3$ functions are near-identical due to the large nonlinear term relative to the linear term.

In practical terms, this is a problem if we want to use a numerical scale that extends beyond $|x_{0} - T| > 1$, but also maintain the same scaling of $d$ in the case of $|x_{0} - T| \leq 1$. For instance, tract variables in the Task Dynamic Application are typically defined over a range of TBCD $\in [-2, 10]$ mm and TBCL $\in [-90\degree, 180\degree]$ \citep{nam-etal2004}. We may wish to use physical measures for simulations, such as tongue tip location in mm, especially when fitting the model to empirical data. One solution is to project the desired scale onto [0, 1], run the simulation, and then project back to the original scale. However, it may also be the case that the relation between movement amplitude and time-to-peak velocity is nonlinear in some regimes but not others, so how are we to capture this fact in order to reproduce the observed characteristics in empirical data? We outline two related solutions below.

\section{Scaling nonlinear terms}
\label{cubic-term-modified}

\subsection{Local scaling for intrinsic movement range}

We begin by normalizing the effects of movement distance on the shape of the velocity trajectory using the inverse square law in (\ref{inv-sq}). An inverse square law holds that a force is inversely proportional to the square of the distance between two masses, as defined by Newton's law of gravitation. In the present case, this has the effect of attenuating the nonlinear term's effect at larger movement amplitudes, such that the effects of nonlinearity are normalized relative to movement distance. Specifically, Equation (\ref{inv-sq}) scales $dk$ by the inverse of the square of the absolute difference between initial position ($x_{0}$) and the target ($T$).  $d$ is bounded in the range $\{d \in \mathbb{R} \ | \ 0 \geq d < 1\}$, where $d$ can be arbitrarily close to 1 given sufficient values of $k$ relative to duration.

\begin{equation}
d' =  \frac{dk}{|x_{0} - T|^2}
\label{inv-sq}
\end{equation}

\begin{figure}[h]
\includegraphics[width=7.5cm]{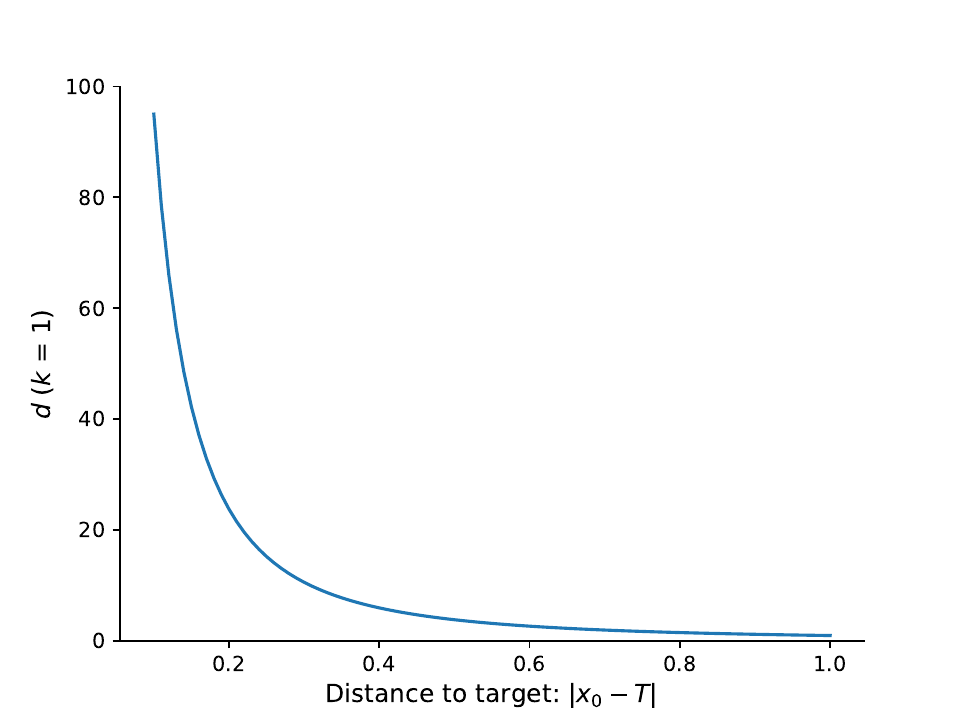}
\includegraphics[width=7.5cm]{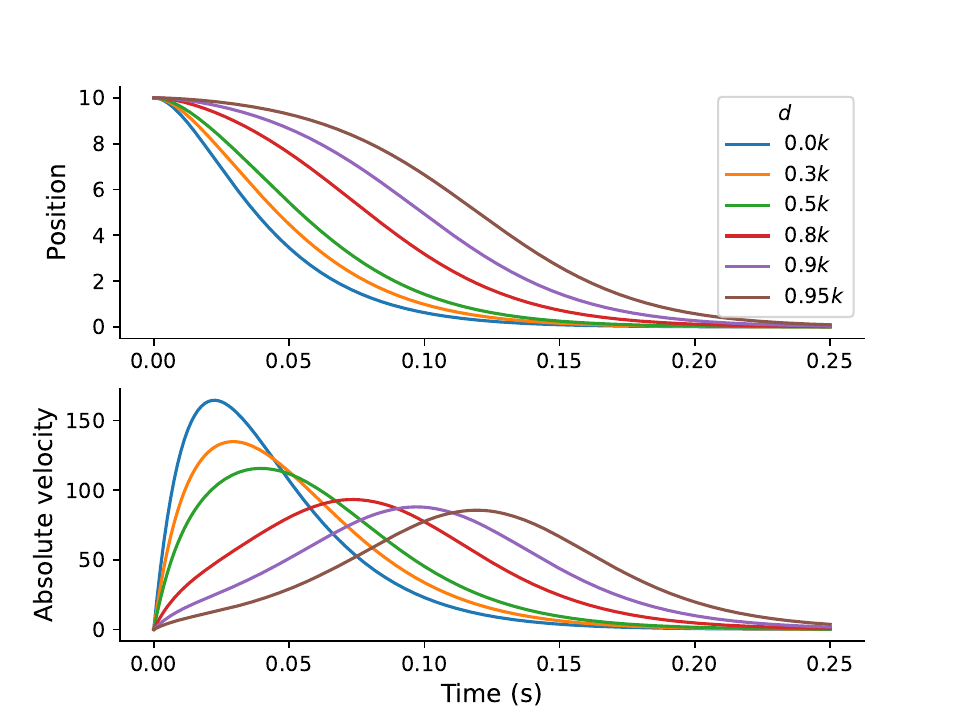}\\
\includegraphics[width=7.5cm]{restoring_force_range10.pdf}
\includegraphics[width=7.5cm]{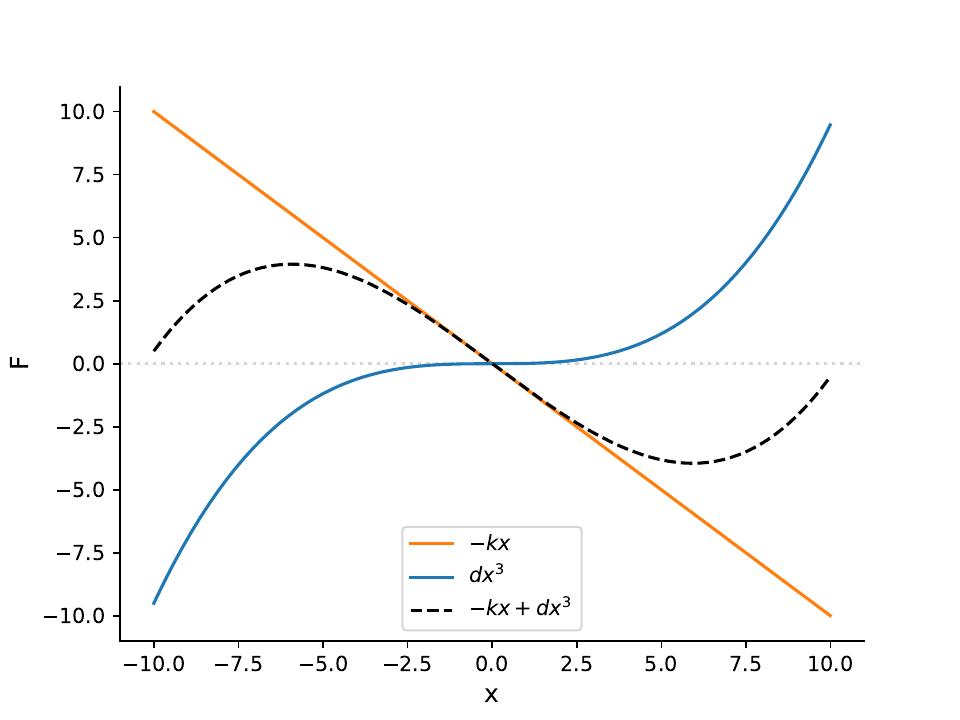}
\caption{TOP LEFT: The relationship between distance-to-target $|x_{0}-T|$ and $d$ follows an inverse square law. TOP RIGHT: The inverse square law allows for appropriate scaling of larger movement distances, with $x_{0} = 10$, $T = 0$, $k = 2000$ across varying values of $d$. BOTTOM LEFT: The restoring forces for $d = 0.95k$ over the range $[-10, 10]$ without scaling. BOTTOM RIGHT: The restoring forces for $d = 0.95k$ over the range $[-10, 10]$ scaled by an inverse square law.}
\label{inverse-square}
\end{figure}

Figure \ref{inverse-square} (top left) shows the required value of $d$ to produce the same time-to-peak velocity across different movement distances between $\{0.1 \geq |x_{0} - T| \leq 1.0\}$, where $d = 0.95$ and $k = 1$.\footnote{The use of $k = 1$ in this plot is simply to visualize the observed relationship over a smaller y-axis range. As $k$ is a constant in the  \citet{sorensen-gafos2016} model, higher values will simply scale the observed relationship accordingly.} The top right panel applies to this a larger movement range, where $x_{0} = 10$ and $T = 0$ across different values of $d$. In this case, the movement range spans $\{0 \geq |x_{0} - T| \leq 10\}$. Scaling each trajectory by its intrinsic $|x_{0} - T|$ reproduces the exact same pattern as the left panel of Figure \ref{fig:d}, preserving the nonlinear relationship between $d$ and time-to-peak velocity, but over a wider parameter range. For this local scaling, we scale by $|x_{0} - T|$ for each trajectory, not the possible movement range across all trajectories. The bottom row in Figure \ref{inverse-square} shows the effects of unscaled and scaled versions of $d$ in terms of the restoring forces. In the left panel, the cubic term dominates and quickly goes to extreme values. In the right panel, the forces are equivalent to those in Figure \ref{models}, but scaled for a range of $x \in [-10, 10]$.

This relation can be generalized for any polynomial term $\alpha x^n$, where $\alpha$ is a scaling coefficient and $n \geq 1$ is the exponent of $x^n$; for example, $\alpha x^1, \alpha x^2, \alpha x^3$, etc. Note that in the case of $\alpha x^1$ the denominator will be raised to the power $1-1 = 0$, where $x^0 = 1$, which means that for linear terms the equation simplifies to $\alpha' = \alpha k$.

\begin{equation}
\alpha' =  \frac{\alpha k}{|x_{0} - T|^{n-1}}
\label{scaling}
\end{equation}

\subsection{Global scaling for potential movement range}

\begin{figure}[h]
\includegraphics[width=7.5cm]{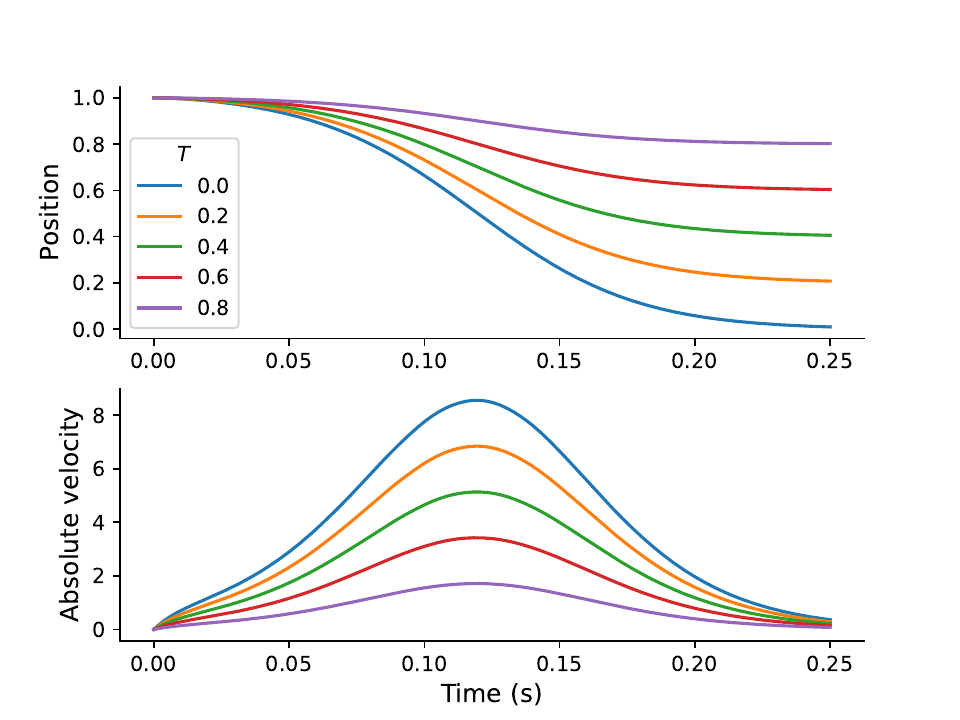}
\includegraphics[width=7.5cm]{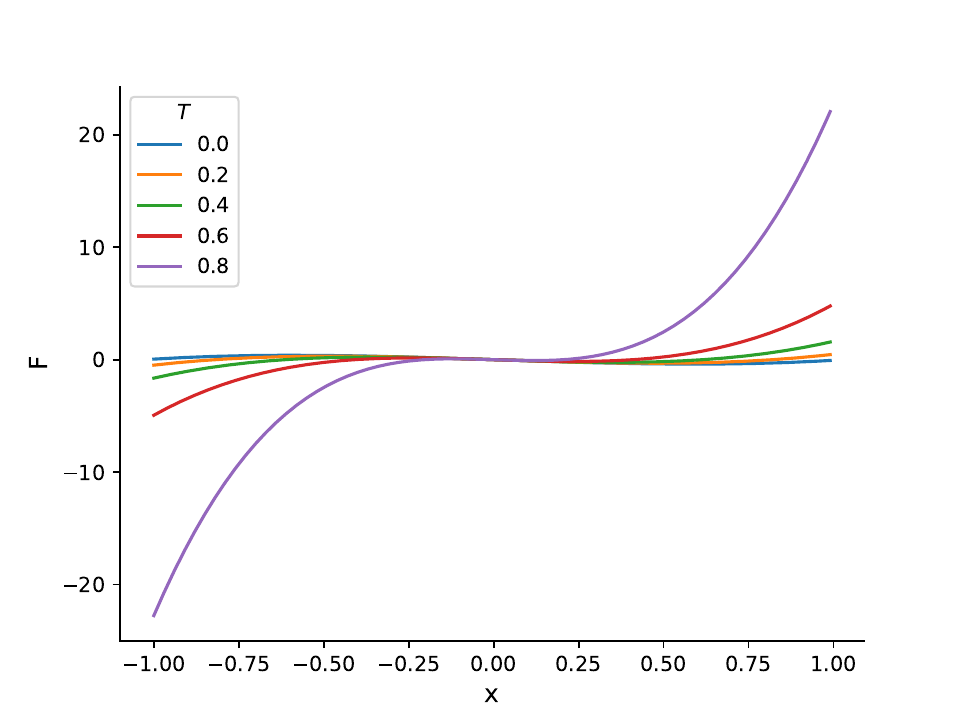}\\
\includegraphics[width=7.5cm]{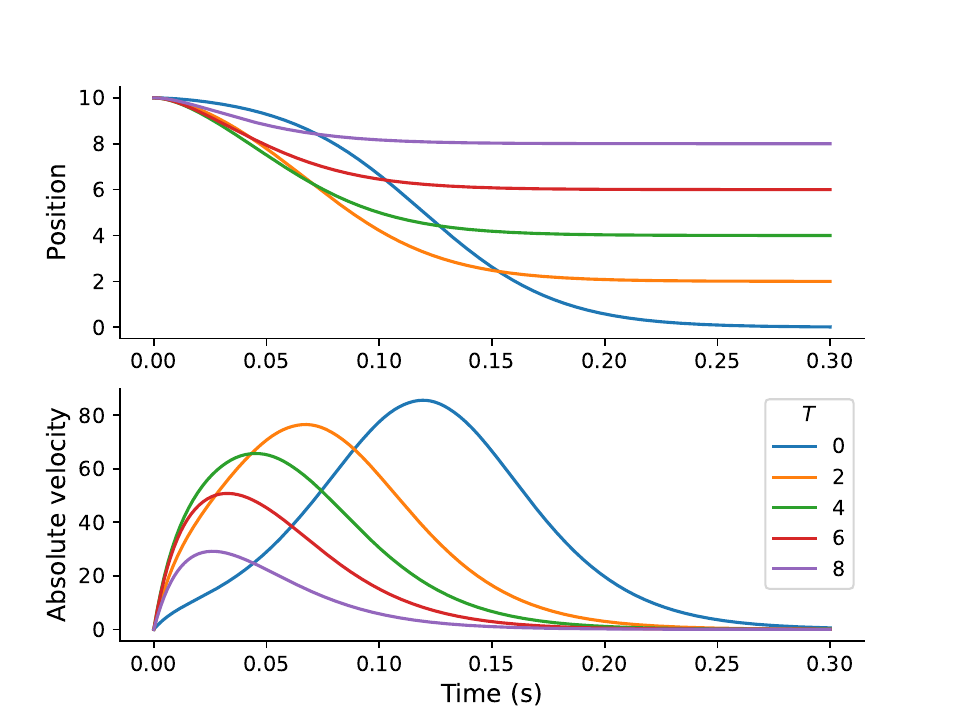}
\includegraphics[width=7.5cm]{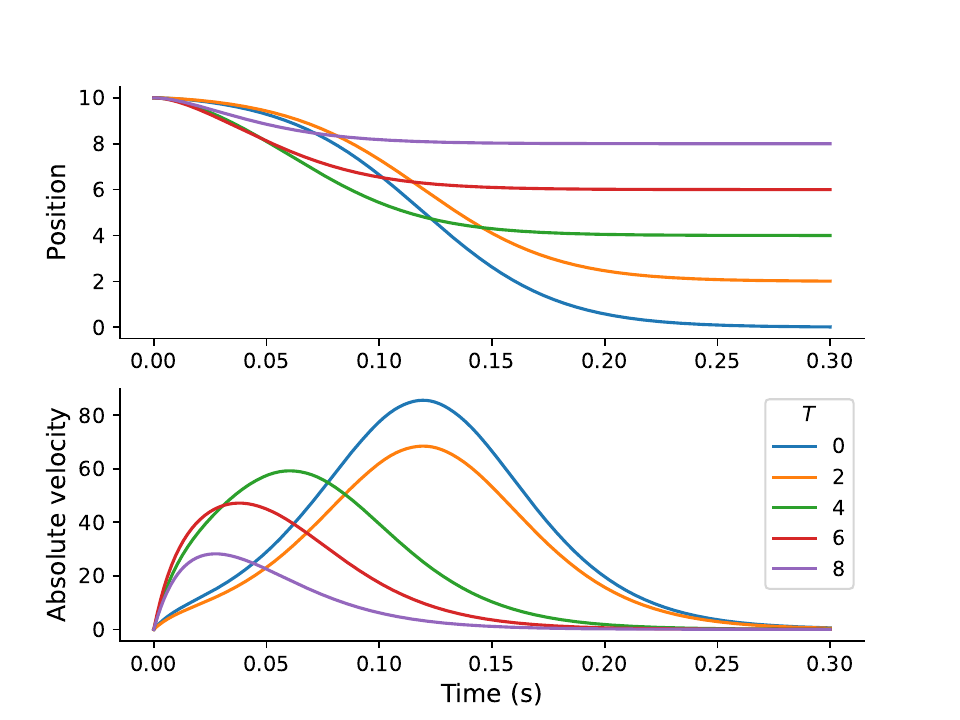}
\caption{TOP LEFT: Cubic model with scaling across different targets in the range $[0, 0.8]$ using an inverse square law. TOP RIGHT: Forces corresponding to the scaled cubic model in top left. BOTTOM LEFT: Cubic model with parameter-range scaling across different targets in the range $[0, 8]$. BOTTOM RIGHT: Cubic model with restricted parameter-range scaling to allow nonlinearity to only operate when $|x_{0}-T| < 8$.}
\label{fig:scaled_plots}
\end{figure}

While the above formulation provides a principled method for normalizing the nonlinear cubic term, it fails to reproduce nonlinear relations between movement amplitude and time-to-peak velocity, thus losing a key feature of the \citet{sorensen-gafos2016} model. For example, Figure \ref{fig:scaled_plots} (top left) shows the effect of $d = 0.95k$ across different movement distances with power law scaling. The corresponding restoring functions $dx^3$ for each movement distance are shown in Figure \ref{fig:scaled_plots} (top right). As a consequence, movement duration is constant and time-to-peak velocity is identical. The only variation is in the amplitude of peak velocity, showing that larger movements involve greater velocities and smaller movements involve smaller velocities. Essentially, this reproduces the dynamics of a linear model across movement distances, but the scaled nonlinear term allows for variation in the shape of the velocity trajectories. To re-state, in this instance, the nonlinear restoring force has been scaled proportionally for each trajectory separately, based on the distance between its initial position and target, but this has eliminated any relationship between movement distance and time-to-peak velocity.
 
We now introduce a small modification to the scaling law, which reintroduces nonlinearity across different movement distances. We first define $D$ as the total \textit{possible} range for a given articulator or tract variable $x'$. Note that $D$ represents the lower and upper bounds of $x'$ across all possible movement trajectories for a given articulatory or tract variable, whereas $|x_{0} - T|$ is the intrinsic movement distance for a particular trajectory.

\begin{equation}
D = |x'_{max} - x'_{min}|
\label{distance}
\end{equation}

We then introduce a scaling factor $\lambda$, which is defined as the ratio between a trajectory's movement range $|x_{0} - T|$ and the total possible movement range $D$. This ratio has an upper bound of 1, as defined in equation (\ref{lambda}).
\begin{equation}
\lambda = \min\left(1, \frac{|x_{0} - T|}{D}\right)
\label{lambda}
\end{equation}

We can therefore add $\lambda$ to the previous generalized Equation (\ref{scaling}) to arrive at Equation (\ref{scaling2}), which allows for scaling the normalized nonlinear coefficient within a global movement range. Figure \ref{fig:scaled_plots} (bottom left) shows the use of the scaling law in Equation (\ref{scaling2}) when $x_{0} \in [0, 10]$ and $T = 1$. In this case, $\alpha = d = 0.95$, $k = 2000$ and $D = 10$ to reflect a possible movement range of 10 units. This restores the nonlinear relation between movement distance and time-to-peak velocity.

\begin{equation}
\alpha' =  \frac{\lambda \alpha k}{|x_{0} - T|^{n-1}}
\label{scaling2}
\end{equation}

The conventional parameterization outlined above defines $D$ as the limits of the potential movement range. In practice, however, it can also be defined as the limit in which nonlinear relations between movement amplitude and time-to-peak velocity are active. For example, imagine our possible movement range is $x \in [0,10]$ and we define $D =$ 8, which is 80\% of the possible movement range. In such a case, when $|x_{0} - T| \geq 8$ then $\lambda = 1$ and all trajectories that meet this condition will have the same time-to-peak velocity, but vary in the amplitude of peak velocity. In contrast, when $|x_{0} - T| < 8$ then $\lambda < 1$ and time-to-peak velocity will vary nonlinearly across trajectories with different movement distances. Figure \ref{fig:scaled_plots} (bottom right) illustrates this example, where $x_{0} = 10$; when $T \in [0, 2]$ time-to-peak velocity is constant and the trajectories only differ in the amplitude of the velocity peak, whereas when $T > 2$ there is a nonlinear relation between distance and time-to-peak velocity. This represents one way of defining the nonlinear relation as operating within a particular part of the movement range. An alternative implementation is to define $\lambda$ nonlinearity across the movement range using a trigonometric function, but we leave the exploration of such possibilities for future research.

\section{Conclusion}

The scaling laws outlined in this article provide simple numerical methods for understanding how nonlinear parameters relate to the intrinsic movement range of a given trajectory, as well as in terms of a potential movement range for a tract variable or articulatory variable. The scaling laws act as principled physical constraints on the nonlinear restoring force across different movement ranges and retain the intrinsic dynamics of the \citet{sorensen-gafos2016} model, without any explicit time-dependence during constant gestural activation. However, the scaled model does introduce some new theoretical questions. First, the local trajectory-intrinsic scaling eliminates the dependency of nonlinearity on initial conditions and linearizes the effect of the cubic term across varying movement distances, which is incompatible with empirical observations of nonlinear relations between movement amplitude and velocity \citep{sorensen-gafos2016}. This motivated a global scaling method that expresses the scope of nonlinearity relative to the potential movement range for an articulator or tract variable, which retains dependence on initial conditions within a restricted scope.

Global scaling effectively bounds nonlinearity at a given movement amplitude threshold, which lends itself to two independent but compatible interpretations: (1) anatomic-motoric constraints; (2) cognitive constraints. The anatomic-motoric interpretation holds that potential movement ranges are inherently bounded by the limits of the vocal tract (e.g. different ranges for lip aperture versus tongue body constriction location), such that this parameter reflects a speaker's proprioceptive knowledge of their vocal tract. The cognitive interpretation holds that the potential movement range represents a window of gestural targets that correspond to a given phonological category. The potential movement range will, therefore, vary between phonological categories, including when categories share the same tract variable. This variability implied by the cognitive view is problematic for a model of invariant phonological targets, but is compatible with dynamical models of speech planning where distributions of targets are defined over neural activation fields \citep{roon-gafos2016, tilsen2019, stern-shaw2023, kirkham-strycharczuk2024}. These two proposals are clearly compatible, because anatomical and cognitive factors both constrain movement dynamics, but it remains possible to commit to an anatomic-motoric interpretation without the cognitive interpretation.


In practical terms, the scaling laws have benefits for simulation, because they allow the simulation of comparable (or identical) velocity profiles across different movement distances. This is particularly useful when simulating dynamics across different articulatory variables that may be on different scales, such as lip aperture vs. tongue dorsum constriction degree vs. tongue dorsum constriction location. If we assume that trajectories across all of these variables tend towards symmetrical velocity profiles then the scaling laws provide a simple and principled way of selecting parameters, without having to hand-tune parameters for each trajectory. The scaling laws also assist with model fitting. When fitting a model to data, we usually aim to minimize an objective function, which typically involves having to define initial estimates for parameters. Given the nonlinear dependence of the cubic coefficient on movement distance, it is challenging to provide initial estimates that are robust to the wide range of movement variation in a data set. This increases the likelihood that the model fails to converge or find an optimal solution. The use of scaled nonlinear coefficients in the target model allows for a much narrower range of estimates, given that $d$ in the cubic model outlined here can only take values between 0 and 1.

The introduction of nonlinear task dynamic models of the speech gesture was a major advance in the development of dynamical theories of articulatory control. Despite this, it is still common for simulation research to use linear models, partly because their parameterization is much simpler, despite the fact that they are often a poor fit with empirical data. The present study demonstrates that the parameterization of nonlinear models can be simplified via scaling laws. The scaling laws also advance the development of dynamical phonological theory by providing physical and cognitive constraints on computational models of articulatory control.

\begin{acknowledgments}
\noindent
Many thanks to the four reviewers for their constructive feedback. Any remaining errors are entirely my own responsibility. This research was supported by UKRI grant AH/Y002822/1.
\end{acknowledgments}

\bibliography{bibliography.bib}

\end{document}